\pdfoutput=1
\documentclass{article}

\PassOptionsToPackage{numbers, compress}{natbib}

\usepackage[main, final]{neurips_2026}

\usepackage[utf8]{inputenc} 
\usepackage[T1]{fontenc}    
\usepackage{hyperref}       
\usepackage{url}            
\usepackage{booktabs}       
\usepackage{amsfonts}       
\usepackage{nicefrac}       
\usepackage{microtype}      
\usepackage{xcolor}         
\usepackage{times}
\usepackage{booktabs}
\usepackage{array}
\usepackage{multirow}
\usepackage{tabularx}
\usepackage{graphicx}
\usepackage{xcolor}
\usepackage{listings}
\usepackage{amsmath}
\usepackage{amssymb}
\usepackage{hyperref}
\usepackage{enumitem}
\usepackage{subcaption}
\usepackage{algorithm}
\usepackage{algpseudocode}
\usepackage{caption}
\usepackage{wrapfig}
\usepackage{needspace}
\usepackage{listings}
\colorlet{revisionblue}{black}
\newcommand{\revision}[1]{{\color{revisionblue}#1}}
\newenvironment{revisionblock}{\begingroup\color{revisionblue}}{\par\endgroup}
\newsavebox{\tabbox}
\title{From History to State: Constant-Context Skill Learning for LLM Agents}

\author{%
  Haoyang Xie, Xinyuan Wang, Yancheng Wang, Puda Zhao, Feng Ju\thanks{Corresponding author} \\\\
  School of Computing and Augmented Intelligence, Arizona State University \\
  \texttt{\{hxie40, xwang735, ywan1053, pzhao34, fengju\}@asu.edu}
}

\begin{document}

\maketitle
\setcounter{footnote}{0}

\begin{abstract}
Large language model (LLM) agents are increasingly used to operate browsers, files, code and tools, making personal assistants a natural deployment target. Yet personal agents face a privacy-cost-capability tension: cloud models execute multi-step workflows well but expose sensitive intermediate context to external APIs, while local models preserve privacy but remain less reliable. Both settings also pay repeatedly for long skill prompts and growing histories. We propose constant-context skill learning, a context-to-weights framework for recurring agent workflows: reusable procedures are learned in lightweight task-family modules, while inference conditions only on the current observation and a compact state block. A deterministic tracker renders this state block from task progress and supplies aligned subgoal rewards, so each module can be trained with step-level SFT and refined through online RL. Across ALFWorld, WebShop, and SciWorld, our agents achieve strong performance across Qwen3-4B, Qwen3-8B\revision{, Qwen3-14B,} and Llama-3.1-8B. With Qwen3-8B, SFT+RL reaches 89.6\% unseen success on ALFWorld, 76.8\% success on WebShop, and 66.4\% unseen success on SciWorld. They match or exceed strong published agent-training results while reducing prompt tokens per turn by 2--7$\times$ relative to controlled ReAct prompting baselines, showing that procedural context can be moved from prompts into weights. \revision{Code is available on \href{https://github.com/haoyangxie/constant-context-skill-learning}{\textcolor{blue}{\underline{GitHub}}}.}
\end{abstract}

\section{Introduction}\label{sec:inteo}
Large language model (LLM) agents are moving from conversational assistants toward systems that execute real multi-turn tasks across browsers, files, code, GUIs, and other tools~\citep{li2025beyond,yao2022webshop,yao2022react,schick2023toolformer,patil2024gorilla,yang2024swe}. As agents become capable of operating tools and files, a natural next step is to deploy them as personal assistants. Systems such as OpenClaw illustrate this direction, where agents interact with personal devices, local tools, and user feedback during normal use~\citep{openclaw2026website}. Such local and user-specific deployment makes privacy and cost first-order concerns. Today, the strongest agent models are often accessed through external APIs~\citep{openai_gpt55_2026,anthropic_claude47_2026}, while models that can run locally are typically less reliable on long-horizon interactive tasks~\citep{team2026qwen3}. As a result, personal agents face an uncomfortable trade-off. Cloud models provide stronger execution, but personal agents may handle sensitive data such as emails, calendars, code, documents, and tool outputs, making it undesirable to send every intermediate state to an external service. At the same time, token costs grow quickly with daily use when every step must process long skill descriptions and accumulated execution histories.

Current approaches still treat agent behavior largely as a context-management problem. ReAct-style agents preserve task progress by replaying prior actions, observations, and reasoning traces in the prompt~\citep{yao2022react}. Memory-augmented agents reduce this burden by storing, organizing, or retrieving past interactions~\citep{zhang2025survey,xu2025mem,shu2026remem,fang2025lightmem}. Recent agent-training methods improve the policy through expert trajectories, self-generated experience, or multi-turn preference and reinforcement-learning objectives~\citep{shi2024direct,fu2025agentrefine,song2024trial,feng2025group,zhang2025agent,liu2026exploratory}. These directions substantially improve long-horizon behavior, but they largely preserve the same inference-time interface: before each action, the agent is still prompted with task instructions, skill descriptions, retrieved memories, demonstrations, or accumulated histories. For recurring personal workflows, this is inefficient: the agent repeatedly re-reads procedures it has already practiced, instead of using that experience as part of the policy itself.

A recurring workflow should not require the agent to reread the same procedure, instructions, and examples at every execution. In personal settings, successful executions naturally accumulate as users repeat the same workflow, and even tens of expert trajectories provide many state-action decisions for supervision. This makes the task family a natural unit of learning. Rather than treating repeated executions as isolated episodes, we group variants of the same skill when they share a tool environment, action space, and procedural subgoals. Learning at this level encourages the module to capture the shared procedure across variants, rather than replaying a fixed action sequence from the training trajectories. The prompt only needs the information that changes from step to step, including the current observation and a compact state block describing progress, relevant entities, and remaining subgoals. Inference is therefore shorter but still well specified. The agent need not reread long skill descriptions or replayed histories, yet it has enough state to choose the next action. We keep the base model fixed and learn a separate lightweight skill module for each task family, first from expert demonstrations and then through interaction. Adding a new workflow only requires training a new module. At inference time, running a workflow loads its corresponding module, keeping computation modest and avoiding catastrophic forgetting across unrelated skills.

\textbf{Our solution}: We instantiate this idea as a context-to-weights training pipeline. For each task family, we train a lightweight skill module with step-level supervised fine-tuning~(SFT) and then refine it with interaction-based reinforcement learning~(RL). A deterministic task tracker provides the interface between raw interaction and learning. It summarizes control-relevant progress into a compact state block, such as acquired objects, completed subgoals, and remaining requirements, and also supplies shaped subgoal rewards during RL. The tracker is a deterministic algorithm, not an LLM reasoner. It converts observations and actions into task progress through simple environment-specific rules. At inference time, the agent loads the corresponding module and conditions on the current observation and state block, rather than the full execution history or long skill prompt. We evaluate this design on ALFWorld, WebShop, and SciWorld~\citep{shridhar2020alfworld, yao2022webshop, wang2022scienceworld}, which provide representative recurring workflows analogous to personal-agent use cases, including household manipulation, web interaction and procedural tool use.

Our experiments show that this context-to-weights design improves both task performance and context efficiency. With Qwen3-8B, SFT+RL reaches 83.6\% / 89.6\% success on ALFWorld seen/unseen, 76.8\% success on WebShop, and 62.9\% / 66.4\% success on SciWorld seen/unseen. At the same time, it substantially reduces inference context. Compared with full-history ReAct, our method reduces prompt tokens per turn from 1.3k to 0.18k on ALFWorld, from 3.1k to 0.49k on WebShop, and from about 2.0k to 0.49k on SciWorld. It also reduces total tokens per episode from 34k to about 3k on ALFWorld, from 47k to 3.4k on WebShop, and from 42k to 18k on SciWorld. Building on these findings,
we summarize our main contributions as follows:

\begin{itemize}
    \item We formulate recurring personal-agent workflows as \emph{constant-context skill learning}, where reusable procedures are moved from prompts and growing histories into lightweight skill-module weights, while step-specific progress remains explicit in a compact state block.
    
    \item We introduce a modular SFT+RL training pipeline that learns lightweight skill modules from expert trajectories and refines them through interaction, pairing deterministic state-block construction with subgoal rewards aligned to the same tracker fields.
    
    \item We conduct controlled evaluations, ablations, and context-efficiency analyses across ALFWorld, WebShop, and SciWorld using Qwen and Llama backbones, showing that bounded-context skill modules can match strong agent-training baselines while reducing prompt-token cost by 2--7$\times$ relative to ReAct baselines.

\end{itemize}

\section{Problem Setup}\label{sec:problem_setup}
\noindent\textbf{Task families.}
We study recurring agent workflows organized as task families. A task family \(k\) is a set of episode variants that share a tool environment, action space, and procedural structure, while differing in episode-specific goals, objects, constraints, or initial states. In personal-agent use, such families arise from repeated workflows, such as ordering items through the same interface, organizing files with the same toolchain, or running variants of the same analysis procedure. In benchmarks, they correspond to predefined families such as ALFWorld object manipulation, WebShop shopping, and SciWorld science-lab tasks. This is a natural unit of learning because variants share reusable procedure but still require episode-specific decisions. At the start of an episode, the agent receives an instruction \(g \sim \mathcal{G}_k\). At step \(t\), it observes \(o_t\), produces textual action \(a_t\), and continues until termination, yielding \(\tau=(g,o_1,a_1,\ldots,o_T,a_T)\). For each family, we assume a modest set of successful executions \(\mathcal{T}_k\), from user demonstrations or prior successful agent runs. \revision{We use the task-family definitions provided by each benchmark. Specifically, these are ALFWorld's six official task types, WebShop's shared shopping workflow, and SciWorld's ETO grouping. We use five ALFWorld adapters because Pick2 reuses the Pick adapter twice.}

\noindent\textbf{Standard history-based formulation.}
Most prompting-based agents maintain task progress by placing more information into the context. Let \(d_k\) denote task-family context such as skill instructions, demonstrations, retrieved memories, or examples, and let \(h_t=(o_1,a_1,\ldots,o_t)\) be the execution history. A standard history-based agent predicts actions from
\(c_t^{\mathrm{hist}}=\mathrm{Format}(g,d_k,h_t)\). This formulation covers ReAct-style prompting, retrieved-memory agents, and skill-prompted agents: they differ in how \(d_k\) or \(h_t\) is assembled, but the next action is still generated from a text context that contains reusable procedure and accumulated interaction state. Its length scales as
\begin{equation}
|c_t^{\mathrm{hist}}|
\approx
|g| + |d_k| + \sum_{i=1}^{t} (|o_i|+|a_i|),
\label{eq:history-growth}
\end{equation}
This is inefficient for recurring workflows. \(d_k\) is reprocessed at every step, while the history term grows with the episode length and can make relevant state harder to use reliably~\citep{liu2024lost}.

\noindent\textbf{Bounded-context skill learning.} We instead seek a task-family policy whose input remains bounded as the episode grows. For each family \(k\), we keep a frozen base model \(\theta_0\) and learn a lightweight skill module \(\phi_k\). At step \(t\), the model conditions on
\(x_t=\mathrm{Format}(g,o_t,q_t,b_t)\), where \(q_t=(o_{t-1},a_{t-1})\) is one-step context and \(b_t\) is a compact state block. The state block contains progress information, such as the current subgoal, selected entities, acquired objects, chosen options, checked locations, or remaining requirements. The desired policy is
\begin{equation}
a_t \sim \pi_{\theta_0,\phi_k}(\cdot \mid x_t),
\qquad
|x_t| \le B_k \quad \forall t,
\label{eq:bounded-policy}
\end{equation}
where \(B_k\) does not scale with the trajectory length. Thus, \(\phi_k\) represents the reusable procedure, while \(x_t\) carries only step-specific state. This bounded-context objective creates a state challenge that the agent must avoid carrying the full history, but it also cannot simply discard history altogether. The current observation alone may omit important progress information, such as selected entities or completed subgoals, and different tasks require different pieces of past state. We use \(b_t\) to denote the compact state block that should carry this information. Its role is not to restate the whole trajectory or provide a solution plan, but to expose the minimal state variables needed for the next decision.

\section{Context-to-Weights Skill Learning}\label{sec:method}

\begin{figure}[t]
    \centering
    \includegraphics[width=\linewidth]{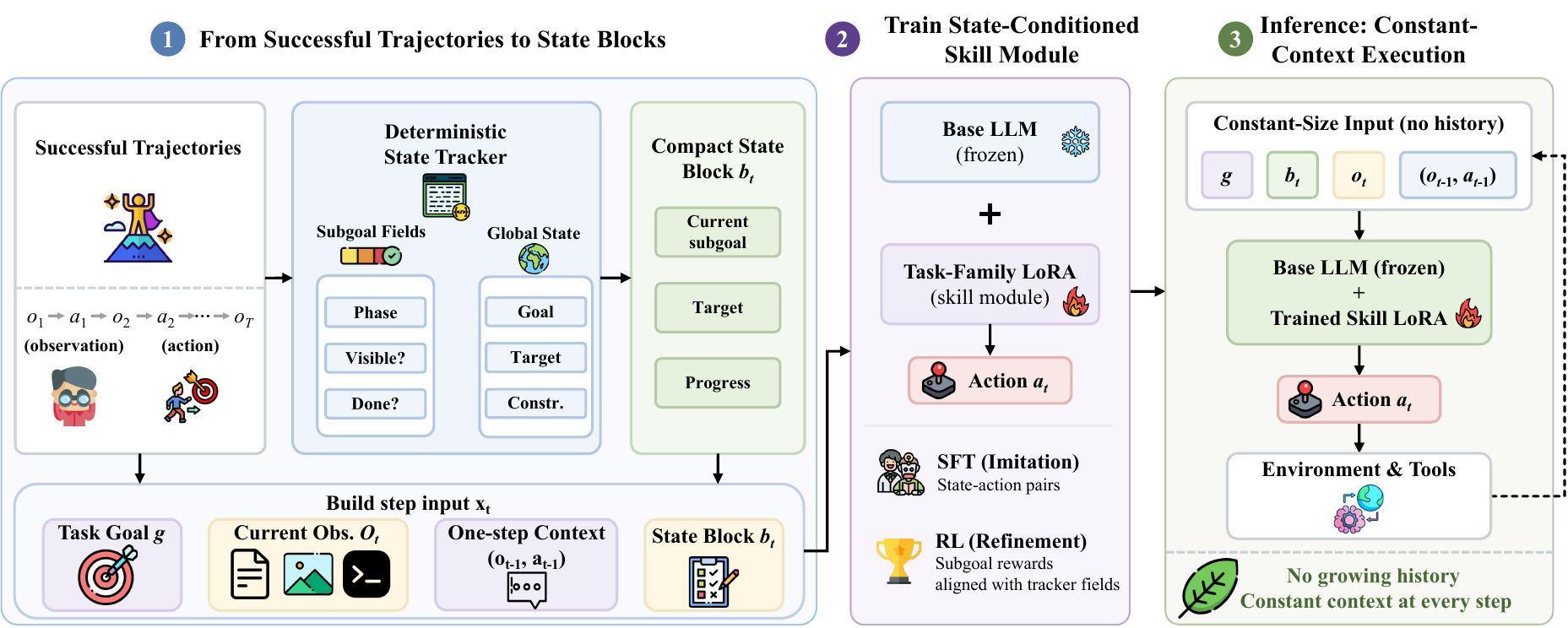}
    \caption{Context-to-weights skill learning pipeline.}
    \label{fig:pipeline}
    \vspace{-15pt}
\end{figure}

Given the bounded-context formulation in Section~\ref{sec:problem_setup}, we now instantiate it as a concrete context-to-weights training pipeline. The pipeline has three stages~(Figure~\ref{fig:pipeline}). First, a deterministic task tracker compresses the relevant interaction state into a compact state block. Second, successful executions are converted into step-level supervision for training a task-family skill module. Third, the module is refined through interaction using rewards derived from the same subgoal structure. Throughout training and inference, the base model \(\theta_0\) remains frozen, only the task-family module \(\phi_k\) is updated.

\subsection{From History to State Blocks}

The purpose of the state block is to preserve decision-relevant progress without passing the full execution history to the model. For each task family \(k\), we define a deterministic tracker \(\mathcal{M}_k\) that maintains a structured state \(m_t\)~\citep{henderson2014word,zhang2020probabilistic}. Initialized from the task instruction \(g\), \(m_t\) stores variables from which the state block can be rendered, including task targets, acquired or selected entities, checked locations, completed subgoals, and other family-specific progress indicators. Within a task family, an episode progresses through procedural phases or subgoals, such as search, manipulation, selection, or placement. The tracker records the current phase as part of \(m_t\). At step \(t\), the tracker updates its state as \(m_t=\mathrm{Update}_k(m_{t-1},a_{t-1},o_t)\) and renders the model-facing state block as \(b_t=\mathrm{Render}_k(m_t)\). Algorithm~\ref{alg:tracker-render} gives the generic update-and-render interface.

The tracker is deterministic rather than an LLM summarizer, so its outputs are reproducible and require no additional model calls. It updates task-family variables with lightweight parsing rules, making state construction inexpensive. The rendered block records only state facts needed for the next decision, not a trajectory summary or full solution plan.

\revision{To reduce manual engineering when extending the framework to new workflows, we also investigate whether an LLM can generate the state schema, tracker update logic, and reward rules offline. Appendix~\ref{app:reward_prompts} provides a representative reward-drafting prompt. We evaluate the generated tracker and reward pipeline on WebShop in Section~\ref{sec:tracker_construction}, with generation and validation details in Appendix~\ref{app:auto_tracker}. The generated tracker runs deterministically without LLM calls during training or inference.}

The tracked variables depend on the family and phase of the task. In ALFWorld, the state block records fields such as the target object, whether the agent is holding it, the destination receptacle, and checked locations. In WebShop, it tracks the current query, inspected product, selected options, remaining options, and purchase readiness. In SciWorld, it tracks the current phase, selected entity, inventory, answer room, and completed subgoals. Across these settings, the same interface is used: the policy receives the current observation \(o_t\), one-step context \(q_t=(o_{t-1},a_{t-1})\), and the rendered state block \(b_t\):
\begin{equation}
x_t = \mathrm{Format}(g,o_t,q_t,b_t).
\label{eq:model-input}
\end{equation}
Thus, past interactions affect the policy only through the tracker state and one-step context, rather than through a growing history window.

\vspace{-0.2cm}
\noindent
\begin{minipage}[t]{0.49\linewidth}
\vspace{0pt}
\begin{algorithm}[H]
\caption{Tracker Update and Render}
\label{alg:tracker-render}
\footnotesize
\begin{algorithmic}[1]
\Require \(k,g,m_{t-1},a_{t-1},o_t\)
\Ensure \(m_t,b_t\)
\State \(z_t \leftarrow \mathrm{Parse}_k(g,o_t,a_{t-1})\)
\Statex \hspace{\algorithmicindent}\emph{entities, locations, options, outcomes}
\State \(m_t \leftarrow m_{t-1}\)
\State Update global fields in \(m_t\) from \(z_t\)
\Statex \hspace{\algorithmicindent}\emph{targets, constraints, inventory}
\State Update phase-specific fields in \(m_t\) from \(z_t\)
\Statex \hspace{\algorithmicindent}\emph{checked locations, selected options, results}
\State \(b_t \leftarrow \mathrm{Render}_k(m_t)\)
\Statex \hspace{\algorithmicindent}\emph{render fields relevant to current phase/subgoal}
\State \Return \(m_t,b_t\)
\end{algorithmic}
\end{algorithm}
\end{minipage}
\hfill
\begin{minipage}[t]{0.49\linewidth}
\vspace{0pt}
\begin{algorithm}[H]
\caption{State-Action Samples}
\label{alg:state-block-construction}
\footnotesize
\begin{algorithmic}[1]
\Require \(k,\mathcal{T}_k,\mathcal{M}_k\)
\Ensure \(\mathcal{D}_k\)
\State \(\mathcal{D}_k \leftarrow \emptyset\)
\For{each \(\tau=(g,o_1,a_1^\star,\ldots,o_T,a_T^\star)\in\mathcal{T}_k\)}
    \State \(m_0 \leftarrow \mathcal{M}_k.\mathrm{Init}(g);\quad o_0,a_0^\star \leftarrow \varnothing\)
    \For{\(t=1,\ldots,T\)}
        \State \(m_t \leftarrow \mathcal{M}_k.\mathrm{Update}(m_{t-1},a_{t-1}^\star,o_t)\)
        \State \(b_t \leftarrow \mathcal{M}_k.\mathrm{Render}(m_t)\)
        \State \(x_t \leftarrow \mathrm{Format}(g,o_t,(o_{t-1},a_{t-1}^\star),b_t)\)
        \State \(\mathcal{D}_k \leftarrow \mathcal{D}_k \cup \{(x_t,a_t^\star)\}\)
    \EndFor
\EndFor
\State \Return \(\mathcal{D}_k\)
\end{algorithmic}
\end{algorithm}
\end{minipage}

\subsection{Training Task-Family Skill Modules}

For each task family \(k\), we replay successful executions from \(\mathcal{T}_k\) through the tracker and collect step-level supervision pairs as detailed in Algorithm~\ref{alg:state-block-construction}, where \(a_t^\star\) denotes the expert action from the successful trajectory. Algorithm~\ref{alg:state-block-construction} invokes the update-and-render procedure in Algorithm~\ref{alg:tracker-render} at each step.
 This yields
\begin{equation}
\mathcal{D}_k
=
\{(x_t,a_t^\star): \tau \in \mathcal{T}_k,\; t=1,\ldots,T_\tau\}.
\label{eq:sft-dataset}
\end{equation}

The module is trained with standard next-action supervision:
\begin{equation}
\mathcal{L}_{\mathrm{SFT}}^{(k)}
=
-\sum_{(x_t,a_t^\star)\in\mathcal{D}_k}
\log p_{\theta_0,\phi_k}(a_t^\star \mid x_t).
\label{eq:sft-loss}
\end{equation}
The target is the executable environment action, not a reasoning trace or a choice among enumerated valid actions. The policy therefore learns to generate the next action text directly from \(x_t\), rather than relying on a per-step action list supplied in the prompt. In our implementation, each task-family module \(\phi_k\) is one LoRA adapter~\citep{hu2022lora}. LoRA keeps each pretrained matrix \(W_0\) fixed and learns a low-rank update:
\begin{equation}
W_k
=
W_0+\Delta W_k,
\qquad
\Delta W_k
=
\frac{\alpha}{r} B_k^{\mathrm{LoRA}} A_k^{\mathrm{LoRA}},
\label{eq:lora-update}
\end{equation}
where \(A_k^{\mathrm{LoRA}}\in\mathbb{R}^{r\times d_{\mathrm{in}}}\), \(B_k^{\mathrm{LoRA}}\in\mathbb{R}^{d_{\mathrm{out}}\times r}\), and \(r\ll \min(d_{\mathrm{in}},d_{\mathrm{out}})\). The superscript distinguishes these LoRA matrices from the context bound \(B_k\) in Eq.~\ref{eq:bounded-policy}. Only these adapter parameters are trained, while the shared base model remains frozen. This gives each task family its own lightweight skill module without overwriting unrelated skills~\citep{houlsby2019parameter,han2024parameter,xu2026parameter}.

This stage transfers repeated procedural knowledge from prompt-time context into task-family parameters. The module learns how to act from the compact state representation across many variants of the same workflow, while episode-specific information remains explicit in \(x_t\).

\subsection{Refining Skill Modules with Subgoal-Guided RL}

Supervised training imitates successful trajectories, but online execution can lead the learned policy into off-trajectory states. We therefore refine each skill module through online interaction. During RL, rollouts use the same bounded input \(x_t\) as in supervised training: after each action, the environment returns a new observation, the tracker updates \(m_t\), and the next state block \(b_t\) is rendered.

The key design is that the same tracker state \(m_t\) used to render \(b_t\) also supplies subgoal-level reward signals, so RL reinforces progress on the variables exposed in the state block. At each step, the reward combines the benchmark environment signal with deterministic progress and error terms:
\begin{equation}
r_t
=
r_t^{\mathrm{env}}
+
r_t^{\mathrm{prog}}
-
r_t^{\mathrm{err}}.
\label{eq:subgoal-reward}
\end{equation}

Here \(r_t^{\mathrm{env}}\) is the environment score or success signal. The progress and error terms, \(r_t^{\mathrm{prog}}\) and \(r_t^{\mathrm{err}}\), are defined around the same subgoals and state-block fields used by the tracker, following the broader idea of shaping sparse task rewards with intermediate progress signals~\citep{ng1999policy}. As a result, RL encourages the policy to use the state block as its execution state. The progress term \(r_t^{\mathrm{prog}}\) rewards actions that advance the current subgoal, while the error term \(r_t^{\mathrm{err}}\) penalizes invalid, repetitive, or phase-inconsistent actions, together with a small step cost. We use a strong reasoning model~(GPT-5.5~\citep{openai_gpt55_2026}) offline to help design these rules from the task specification and the state-block schema. The rewards are then fully deterministic at rollout time, with no LLM judge queried during RL. Appendix~\ref{app:reward_details} gives the reward specification and Appendix~\ref{app:reward_prompts} shows the prompt used to ask the reasoning model to draft deterministic reward rules from the state-block schema.

We use a GRPO-style group-normalized policy-gradient update~\citep{shao2024deepseekmath}. For each task instance, the current policy samples a group of \(K\) rollouts, indexed by \(i\). Let \(T_i\) be the length of rollout \(i\), and let \(r_u^{(i)}\) be its shaped reward at step \(u\). We compute the discounted step return \(G_t^{(i)}=\sum_{u=t}^{T_i}\gamma^{u-t}r_u^{(i)}\), then normalize these returns across all steps from the \(K\) rollouts of the same task instance:
\begin{equation}
A_t^{(i)}
=
\frac{G_t^{(i)}-\mu_{\mathrm{group}}}
{\sigma_{\mathrm{group}}+\epsilon},
\qquad
\mu_{\mathrm{group}},\sigma_{\mathrm{group}}
=
\mathrm{MeanStd}\left(\{G_t^{(i)}\}_{i,t}\right).
\label{eq:group-advantage}
\end{equation}
Here \(\gamma\) is the discount factor and \(\epsilon\) is a small constant for numerical stability. This compares attempts under the same task difficulty and avoids training a separate value function.

The sampled action at step \(t\) of rollout \(i\) is \(a_t^{(i)}\), generated from the bounded input \(x_t^{(i)}\). We update only the task-family module with a sampled policy-gradient loss and a reference-policy penalty to the frozen SFT adapter:
\begin{equation}
\mathcal{L}_{\mathrm{RL}}^{(k)}
=
\mathbb{E}_{i,t}
\left[
- A_t^{(i)}
\log \pi_{\theta_0,\phi_k}(a_t^{(i)} \mid x_t^{(i)})
+
\beta
\left(
\log \pi_{\theta_0,\phi_k}(a_t^{(i)} \mid x_t^{(i)})
-
\log \pi_{\theta_0,\phi_k^{\mathrm{SFT}}}(a_t^{(i)} \mid x_t^{(i)})
\right)
\right].
\label{eq:rl-loss}
\end{equation}

Here \(\phi_k^{\mathrm{SFT}}\) is the frozen SFT-stage adapter, which keeps RL close to the action format and procedural behavior learned during supervised training.
 During refinement, \(\theta_0\) remains frozen and only \(\phi_k\) is updated.

\begin{table}[!t]
\centering
\caption{Main task performance (mean$\pm$std over inference seeds; \textbf{bold} = best per column). Score is benchmark-provided. SR denotes success rate, measured as binary full-task success (\%).}
\vspace{2pt}
\label{tab:main-results}
\footnotesize
\begingroup
\newcommand{\pmz}[2]{#1{\scriptsize$\pm$#2}}
\setlength{\tabcolsep}{1.5pt}
\begin{tabular*}{\linewidth}{@{\extracolsep{\fill}}>{\centering\arraybackslash}p{1.55cm} >{\centering\arraybackslash}p{1.95cm} c c c c c c c c@{}}
\toprule
\multirow{3}{*}{Base Model} & \multirow{3}{*}{Method} & \multicolumn{2}{c}{ALFWorld} & \multicolumn{2}{c}{WebShop} & \multicolumn{4}{c}{SciWorld} \\
\cmidrule(lr){3-4} \cmidrule(lr){5-6} \cmidrule(lr){7-10}
& & \multicolumn{2}{c}{SR} & \multicolumn{2}{c}{} & \multicolumn{2}{c}{Seen} & \multicolumn{2}{c}{Unseen} \\
\cmidrule(lr){7-8} \cmidrule(lr){9-10}
& & Seen & Unseen & Score & SR & Score & SR & Score & SR \\
\midrule
\multirow{2}{*}{Qwen3-4B}
& Ours: SFT only    & \pmz{59.5}{3.5} & \pmz{56.7}{3.3} & \pmz{66.4}{1.4} & \pmz{59.6}{2.3} & \pmz{67.8}{1.9} & \pmz{59.3}{1.7} & \pmz{61.0}{1.0} & \pmz{52.0}{0.3} \\

& Ours: SFT + RL   & \pmz{76.4}{1.5} & \pmz{81.3}{1.9} & \pmz{86.0}{0.6} & \pmz{76.1}{1.1} & \pmz{72.4}{1.9} & \pmz{63.1}{1.5} & \pmz{75.1}{1.4} & \pmz{65.4}{1.4} \\
\midrule
\multirow{2}{*}{Qwen3-8B}
& Ours: SFT only                        & \pmz{62.4}{2.5} & \pmz{61.7}{2.2} & \pmz{69.5}{1.7} & \pmz{62.2}{1.2} & \pmz{65.7}{1.6} & \pmz{58.2}{1.5} & \pmz{61.5}{0.9} & \pmz{52.0}{0.7} \\
& Ours: SFT + RL        & \textbf{\pmz{83.6}{1.8}} & \textbf{\pmz{89.6}{0.4}} & \pmz{84.0}{0.0} & \pmz{76.8}{0.2} & \pmz{72.8}{2.7} & \pmz{62.9}{2.2} & \textbf{\pmz{79.7}{0.2}} & \pmz{66.4}{0.5} \\
\midrule
\multirow{2}{*}{Llama-3.1-8B}
& Ours: SFT only                        & \pmz{61.0}{3.4} & \pmz{60.2}{1.7} & \pmz{69.1}{1.7} & \pmz{60.3}{1.0} & \pmz{66.5}{1.5} & \pmz{61.0}{1.6} & \pmz{57.6}{0.7} & \pmz{50.6}{1.0} \\

& Ours: SFT + RL                        & \pmz{73.6}{0.7} & \pmz{80.6}{2.6} & \textbf{\pmz{86.3}{1.5}} & \pmz{75.7}{1.3} & \pmz{74.1}{1.0} & \pmz{66.2}{2.1} & \pmz{73.9}{0.6} & \pmz{64.1}{0.3} \\
\midrule
\multirow{2}{*}{Qwen3-14B}
& Ours: SFT only
& \pmz{62.9}{2.7} & \pmz{61.4}{2.3}
& \pmz{66.5}{1.5} & \pmz{59.8}{1.8}
& \pmz{68.9}{1.4}
& \pmz{61.9}{1.3}
& \pmz{61.6}{0.6}
& \pmz{53.1}{0.8} \\

& Ours: SFT + RL
& \pmz{82.1}{0.8} & \pmz{87.3}{1.6}
& \pmz{85.9}{0.5} & \textbf{\pmz{78.8}{0.8}}
& \textbf{\pmz{76.0}{1.6}} & \textbf{\pmz{67.0}{1.8}}
& \pmz{79.1}{0.1} & \textbf{\pmz{67.3}{0.3}} \\

\bottomrule
\end{tabular*}
\endgroup
\vspace{-0.5cm}
\end{table}

\section{Experiments and Results}\label{sec:results}

We evaluate the proposed context-to-weights pipeline on three representative agent benchmarks, using Qwen and Llama backbones at different model scales. Our experiments are organized around four research questions. (\textbf{RQ1}) Does task-family skill learning improve agent performance across benchmarks and backbones, and how much does interaction-based RL refinement add over SFT-only skill modules? (\textbf{RQ2}) How much inference context does the bounded-context interface save compared with ReAct-style prompting, a representative history-based agent interface that repeatedly conditions on long instructions and growing execution histories? (\textbf{RQ3}) Which components of the pipeline, including the state block, one-step context, SFT training, and subgoal-guided reward shaping, are responsible for the gains? (\textbf{RQ4}) How does the method compare with prior agent methods across memory-based prompting, learned memory and RL-based agent training? \revision{We further examine robustness across task variants and the feasibility of LLM-generated trackers and reward rules.}

\subsection{Experimental Setup}\label{sec:experimental_setup}

\begin{figure}[ht]
    \centering
    \captionsetup{skip=2pt}
    \includegraphics[width=1\linewidth]{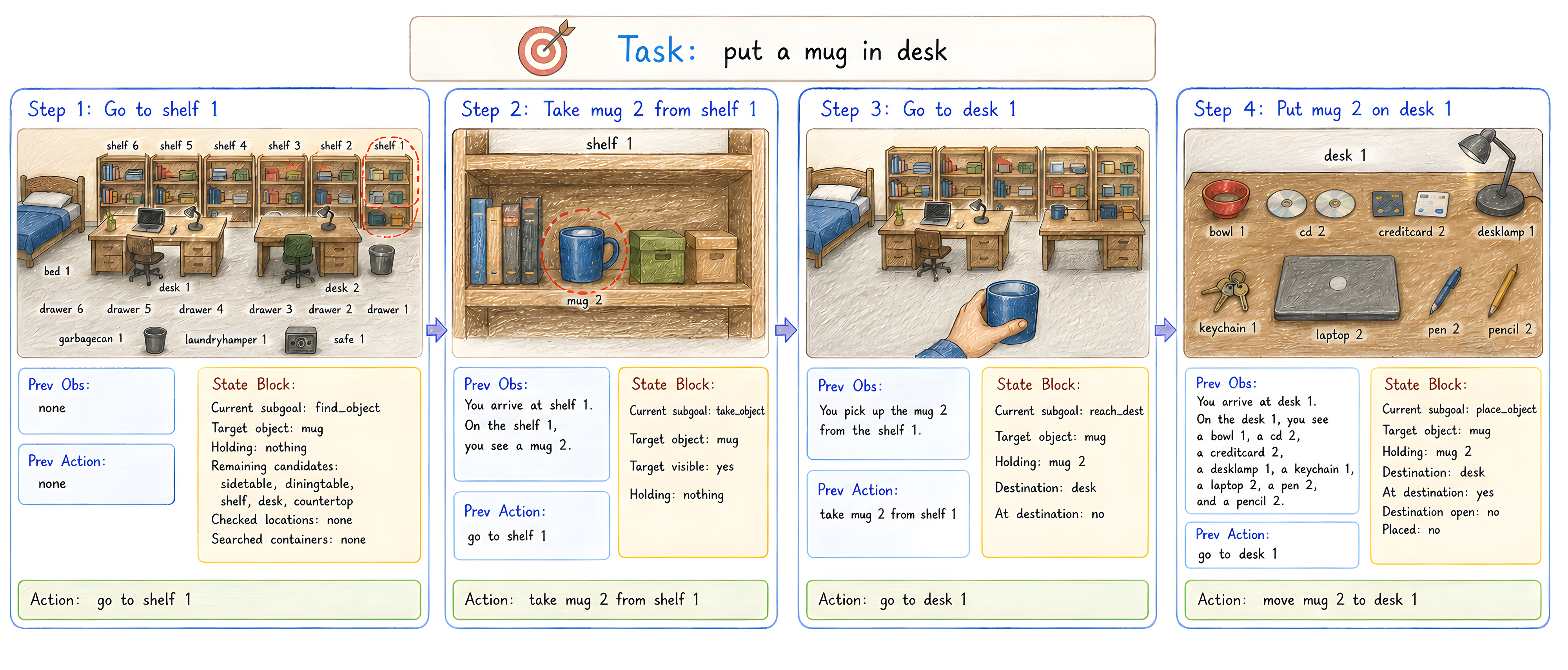}
    \caption{Qualitative ALFWorld household trace with rendered state blocks.}
    \label{fig:alfworld_example}
    \vspace{-0.45cm}
\end{figure}

\noindent\textbf{Benchmarks.}
We evaluate on three text-mode agent benchmarks that approximate recurring personal-agent workflows in different fixed environments. \textbf{ALFWorld}~\citep{shridhar2020alfworld} represents household assistance: the agent navigates simulated rooms, finds target objects, applies transformations such as cleaning, heating, or cooling and places objects in target receptacles. We evaluate on 140 seen and 134 unseen games. \textbf{WebShop}~\citep{yao2022webshop} represents shopping assistance: the agent searches for products, opens item pages, selects required options, and purchases the target item. We use the 1{,}000-product text setting and evaluate on 500 test goals. \textbf{SciWorld}~\citep{wang2022scienceworld} represents procedural tool use: the agent manipulates objects, uses instruments, performs experiments, and reports answers in science-lab environments. We evaluate on 194 seen and 211 unseen variants following the ETO split structure~\citep{song2024trial}. All three use the same agent interface, text observations in and executable text actions out. Figures~\ref{fig:alfworld_example}, \ref{fig:webshop_example}, and~\ref{fig:sciworld_example} show representative interactions from these settings. Additional details on task families, training trajectories, and evaluation settings are provided in Appendix~\ref{app:training_details}.

\noindent\textbf{Baselines.}
Our baselines serve two purposes. For a \emph{controlled token-cost comparison}, we run two ReAct variants under the same evaluation setup as our method: \textbf{ReAct-1step} retains only the most recent observation and action, while \textbf{ReAct-full} appends the available trajectory history with truncation when needed. Prior prompt- and memory-based agents differ in their prompts, retrieved content, and history policies, but they usually include at least recent interaction context and often much more. The two ReAct variants therefore represent the main short-context and history-based settings used by prompting-based agents, giving a controlled efficiency comparison in Table~\ref{tab:context-efficiency}. For \emph{performance comparison}, we report published state-of-the-art agent-training results in Table~\ref{tab:published-comparison}.

\noindent\textbf{Implementation details.}
We use Qwen3-8B as the default backbone and repeat the main experiments with Qwen3-4B\revision{, Qwen3-14B,} and Llama-3.1-8B to test robustness across model scale and family. For each task family, the base model is frozen and only a LoRA skill module is trained, updating about 2\% of the parameters, $0.5$--$0.7$ GB on disk per adapter. We first train modules with SFT and then refine them with online GRPO-style RL using four rollouts per task instance and the subgoal rewards. All main training runs fit on a single NVIDIA A100 80GB GPU. Main performance is reported as mean$\pm$std over three inference seeds with temperature 0.4 and top-$p$ 0.95, while token-cost results use deterministic decoding for stable accounting. Full hyperparameters, decoding settings, and compute details are reported in Appendix~\ref{app:training_details}.

\begin{table}[!t]
\centering
\caption{Context and compute efficiency with Qwen3-8B under controlled prompting comparisons. All methods use the same tokenizer, environment split, and action parser.}
\label{tab:context-efficiency}
\small
\setlength{\tabcolsep}{4pt}
\begin{tabularx}{\linewidth}{@{}l X c c c c c@{}}
\toprule
\multirow{2}{*}{Metric} & \multirow{2}{*}{Method}
& \multicolumn{2}{c}{ALFWorld}
& \multicolumn{1}{c}{WebShop}
& \multicolumn{2}{c}{SciWorld} \\
\cmidrule(lr){3-4} \cmidrule(lr){5-5} \cmidrule(lr){6-7}
& & Seen & Unseen & Test & Seen & Unseen \\
\midrule

\multirow{3}{*}{Avg. Steps}
& ReAct, 1-step history & 24.9 & 25.3 & 13.6 & 18.9 & 12.3 \\
& ReAct, full history   & 21.6 & 21.8 & 14.4 & 13.4 & 12.5 \\
& Ours: SFT + RL        & 18.4 & 15.9 & 6.8 & 33.7 & 30.7 \\
\midrule

\multirow{3}{*}{Prompt Tok./Turn}
& ReAct, 1-step history & 380.0 & 358.6 & 1{,}059.0 & 1{,}443.0 & 1{,}481.0 \\
& ReAct, full history   & 1{,}310.5 & 1{,}320.4 & 3{,}092.7 & 2{,}038.9 & 1{,}938.0 \\
& Ours: SFT + RL        & 183.8 & 179.9 & 488.0 & 492.0 & 496.3 \\
\midrule

\multirow{3}{*}{Completion Tok./Turn}
& ReAct, 1-step history & 53.3 & 52.3 & 72.4 & 61.9 & 65.3 \\
& ReAct, full history   & 50.9 & 52.9 & 71.4 & 54.6 & 54.3 \\
& Ours: SFT + RL        & 7.1 & 7.1 & 10.4 & 5.3 & 5.3 \\
\midrule

\multirow{3}{*}{Total Tok./Episode}
& ReAct, 1-step history & 10{,}969 & 10{,}443 & 15{,}763 & 28{,}088 & 18{,}953 \\
& ReAct, full history   & 34{,}199 & 34{,}145 & 46{,}531 & 42{,}280 & 39{,}550 \\
& Ours: SFT + RL        & 3{,}251 & 2{,}851 & 3{,}353 & 17{,}894 & 16{,}182 \\

\bottomrule
\end{tabularx}
\vspace{-0.4cm}
\end{table}

\subsection{Main Task Performance}\label{sec:task_performance}
Table~\ref{tab:main-results} shows that moving recurring procedures from context into task-family modules is effective across benchmarks. SFT already gives strong performance, indicating that expert trajectories provide enough step-level supervision for the model to act from compact per-step inputs rather than long skill prompts or full histories. On WebShop, for example, the untrained prompt variant with a state block reaches only 23.6\% success in Table~\ref{tab:ablations}, while SFT raises success to 62.2\% with Qwen3-8B. RL refinement further improves the trained modules. With Qwen3-8B, SFT+RL reaches 83.6\% / 89.6\% success on ALFWorld seen/unseen, 76.8\% success on WebShop, and 62.9\% / 66.4\% success on SciWorld seen/unseen. Across the \revision{four} backbones, RL improves over SFT by roughly 13--28 percentage points on ALFWorld, \revision{15--19} points on WebShop, and 5--15 points on SciWorld. The improvements are also not tied to a single backbone. Qwen3-8B gives the strongest ALFWorld performance and the best SciWorld unseen \revision{score}, while Llama-3.1-8B is strongest on WebShop score. \revision{Qwen3-14B reaches 78.8\% WebShop success and 67.0\% / 67.3\% SciWorld seen/unseen success, confirming that the training gains persist at the larger scale.} Qwen3-4B also benefits from the same training recipe despite its smaller size. This pattern suggests that the gains come from the task-family training setup, not from a particular pretrained model. As a modularity check, we also train a single unified Qwen3-8B module across all ALFWorld families, which reaches 74.3\% / 70.9\% seen/unseen success after SFT and 86.4\% / 88.1\% after RL. This shows that a unified module is viable, but we use task-family modules for realistic deployment because they reduce interference among different skills, especially for smaller local models, and avoid training bias from imbalanced task-family data. \revision{A cross-benchmark SFT experiment further shows that a single adapter can support all three environments, with some ALFWorld unseen interference (Appendix~\ref{app:unified_cross_benchmark}).} We analyze the individual contributions of the state block, SFT, and RL in Section~\ref{sec:ablations}.

\subsection{Context and Token Efficiency}\label{sec:token_efficiency}

\revision{The efficiency measurements below concern inference-time tokens. Task quality is reported alongside the training stages in Table~\ref{tab:main-results}, while Table~\ref{tab:published-comparison} includes the controlled ReAct baselines. The token reductions do not include the one-time offline training cost, reported separately in Appendix~\ref{app:training_details}.}

Table~\ref{tab:context-efficiency} shows that the bounded-context interface substantially reduces inference context. ReAct-1step is already a strong efficiency baseline because it keeps only the latest observation and action. Many prompt- and memory-based agents keep more history, retrieved memories, or both.  Our method still uses 2.1--3.0$\times$ fewer prompt tokens per turn than ReAct-1step on all three benchmarks and reduces prompt tokens by about 4--7$\times$ relative to ReAct-full. This saving comes from moving reusable procedural context into the learned module, not merely from dropping history. 

The episode-level totals show the same pattern. On ALFWorld and WebShop, failed executions often continue producing unproductive actions until the step limit, so better execution also reduces average steps. SciWorld behaves differently, invalid actions can terminate an episode early with negative score, while successful procedures can be long, with task-specific limits up to 100 steps. Thus, stronger SciWorld agents may take more steps because they survive and complete more procedures. Even there, our method substantially reduces total token use. Relative to ReAct-full, total token use drops by about 10--12$\times$ on ALFWorld, 14$\times$ on WebShop, and 2--3$\times$ on SciWorld.

\subsection{Component Ablations}\label{sec:ablations}

\noindent
\begin{minipage}[t]{0.38\linewidth}
\vspace{-10pt}
\sbox{\tabbox}{\footnotesize\setlength{\tabcolsep}{3.5pt}\begin{tabular}{@{}l c c@{}}
\toprule
Method & Score & SR \\
\midrule
\multicolumn{3}{l}{\emph{Prompt design (no training):}} \\
Current observation only & 10.1 & 1.2\% \\
+ one-step context & 34.2 & 5.6\% \\
+ subgoal state block & 42.6 & 23.6\% \\
\midrule
\multicolumn{3}{l}{\emph{Trained leave-one-out variants:}} \\
w/o state block & 72.9 & 61.2\% \\
w/o reward shaping & 83.1 & 72.2\% \\
\midrule
\multicolumn{3}{l}{\emph{Trained methods:}} \\
SFT only & 69.5 & 62.2\% \\
Full method & \textbf{84.0} & \textbf{76.8\%} \\
\bottomrule
\end{tabular}}%
\begin{minipage}[t]{\wd\tabbox}
\captionsetup{type=table,width=\linewidth,justification=justified,singlelinecheck=false}
\captionof{table}{Ablation study on \mbox{WebShop} with Qwen3-8B.}
\label{tab:ablations}
\usebox{\tabbox}
\end{minipage}
\end{minipage}
\hspace{-10pt}
\begin{minipage}[t]{0.62\linewidth}
\vspace{-10pt}

We ablate components on WebShop with Qwen3-8B, evaluated on all 500 test goals. This default-backbone,single-module setting makes
component changes easy to isolate.
Table~\ref{tab:ablations} shows that the state block is the main prompt-side signal. Without training, the current observation alone is near floor performance, and adding one-step context only raises success from 1.2\% to 5.6\%. Adding the state block raises success to 23.6\%, a much larger gain, showing that compact progress state is the key information missing from short prompts.

Training then moves reusable procedure into the skill module. SFT raises success from 23.6\% to 62.2\%, and RL further improves it to 76.8\%. The trained leave-one-out variants are all evaluated after RL refinement. Removing the state block drops success by 15.6 points, confirming that the model still uses it as execution state. \revision{Section~\ref{sec:generalization} separately tests robustness to changes in instruction wording, goal combinations, and state serialization.}
\end{minipage}

\Needspace{10\baselineskip}
\begin{wrapfigure}[8]{r}{0.36\linewidth}
    \vspace{-10pt}
    \centering
    \includegraphics[width=\linewidth]{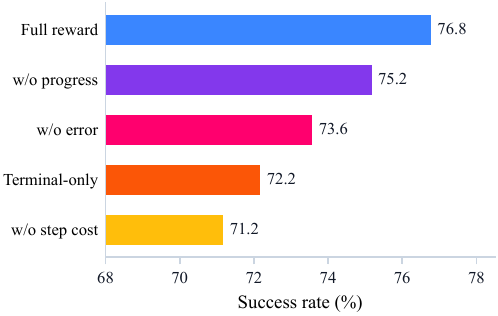}
    \vspace{-8pt}
    \caption{Reward ablation.}
    \label{fig:webshop_reward_ablation}
    \vspace{-12pt}
\end{wrapfigure}

Using only terminal task reward during RL drops success by 4.6 points, showing that subgoal rewards provide useful intermediate feedback. Figure~\ref{fig:webshop_reward_ablation} further decomposes the reward design. Progress rewards guide option and subgoal completion, error penalties reduce invalid or repeated actions, and the step cost discourages unnecessary actions. Removing any component lowers success, showing that the shaped reward works as a combined signal. The full reward specification is listed in Appendix~\ref{app:reward_details}. Appendix~\ref{app:data_efficiency} further studies SFT data \\ efficiency on WebShop.

\begin{revisionblock}
\subsection{Generalization Across Task Variants}\label{sec:generalization}
\setlength{\intextsep}{0pt}
\begin{wraptable}{r}{0.54\linewidth}
\captionsetup{width=\linewidth,labelfont={color=revisionblue},textfont={color=revisionblue},justification=raggedright,singlelinecheck=false}
\caption{Qwen3-8B SFT+RL generalization: SR (\%), mean$\pm$std over three inference seeds; $N$ goals per seed.}
\label{tab:generalization}
\footnotesize
\setlength{\tabcolsep}{3pt}
\begin{tabularx}{\linewidth}{@{}X r c@{}}
\toprule
Input condition & $N$ & SR \\
\midrule
\textit{WebShop}: original & 500 & $76.8\pm0.2$ \\
Paraphrased instruction & 500 & $72.2\pm0.7$ \\
Unseen constraint combination & 446 & $75.8\pm0.6$ \\
State as sorted JSON & 500 & $75.7\pm0.8$ \\
\midrule
\textit{ALFWorld unseen}: original & 134 & $89.6\pm0.4$ \\
Paraphrased instruction & 134 & $89.1\pm1.1$ \\
\bottomrule
\end{tabularx}
\end{wraptable}
To examine robustness across task variants, we test whether the learned policies remain effective under changes to instruction wording, goal composition, and state serialization. We evaluate fixed Qwen3-8B SFT+RL adapters on paraphrased instructions, unseen WebShop constraint combinations, and JSON state serialization (Table~\ref{tab:generalization}). Paraphrases change only model-visible instructions. Tracker and environment goals remain unchanged. Results use three inference seeds (Appendix~\ref{app:generalization_protocol}). State reserialization and ALFWorld paraphrasing retain near-original performance, while unseen constraint combinations achieve 75.8\% SR. This pattern supports robust execution across input variants. WebShop paraphrasing retains 72.2\% SR, indicating greater wording sensitivity.
\end{revisionblock}

\begin{revisionblock}
\subsection{Automating the State Interface}\label{sec:tracker_construction}
\setlength{\intextsep}{0pt}
\sbox{\tabbox}{\footnotesize\setlength{\tabcolsep}{6pt}\begin{tabular}{@{}l c c@{}}
\toprule
Tracker & SFT & SFT+RL \\
\midrule
Manual & 62.2 & 76.8 \\
LLM-generated & 61.2 & 65.8 \\
\bottomrule
\end{tabular}}%
\begin{wraptable}{r}{\wd\tabbox}
\captionsetup{width=\linewidth,labelfont={color=revisionblue},textfont={color=revisionblue},justification=justified,singlelinecheck=false}
\caption{WebShop SR (\%) with Qwen3-8B on 500 goals.}
\label{tab:tracker_construction}
\usebox{\tabbox}
\end{wraptable}
To reduce manual engineering for new workflows, we investigate LLM-based generation of state schemas, trackers, and reward rules. GPT-5.4 generates a WebShop state schema, tracker, renderer, and deterministic reward rules from the public interface and expert training examples, with our tracker and 500 test goals withheld. Replay validation precedes SFT and RL. The generated code requires no runtime LLM calls. SFT with the generated interface reaches 61.2\% SR, close to the manual tracker's 62.2\%. RL further improves it to 65.8\% (Table~\ref{tab:tracker_construction}), demonstrating that the generated state and reward rules support both training stages. Appendix~\ref{app:auto_tracker} gives the full protocol and metrics.
\end{revisionblock}

\subsection{Comparison with Published Methods}\label{sec:comparison}

\noindent
\begin{minipage}[t]{0.42\linewidth}
\vspace{-10pt}
\sbox{\tabbox}{\scriptsize\setlength{\tabcolsep}{3pt}\begin{tabular}{@{}l c c c@{}}
\toprule
Method & ALFWorld & WebShop & SciWorld \\
\midrule
ReAct-1 & 23.1\% & 8.8 / 1.2\% & -50.0 \\
ReAct-full & 46.3\% & 7.0 / 3.8\% & -21.5 \\
ETO~\citep{song2024trial} & 72.4\% & 67.4 / 37.5\% & 65.0 \\
DMPO~\citep{shi2024direct} & 88.8\% & 64.3 / -- & 65.1 \\
GiGPO~\citep{feng2025group} & \textbf{90.8\%} & 84.4 / 72.8\% & -- \\
MemGen~\citep{zhang2025memgen} & 90.6\% & -- & -- \\
Early Exp.~\citep{zhang2025agent} & 85.9\% & 77.0 / 62.2\% & 68.0 \\
EMPO2~\citep{liu2026exploratory} & -- & \textbf{88.3 / 76.9\%} & 75.9 \\
ACT~\citep{liu2026agentic} & 88.1\% & -- / 33.8\% & 50.34 \\
Ours & 89.6\% & 84.0 / 76.8\% & \textbf{79.7} \\
\bottomrule
\end{tabular}}%
\begin{minipage}[t]{\wd\tabbox}
\captionsetup{type=table,width=\linewidth,justification=justified,singlelinecheck=false}
\captionof{table}{Comparison with published baselines. \% denote success rates, otherwise values are task scores.}
\label{tab:published-comparison}
\usebox{\tabbox}
\end{minipage}
\end{minipage}
\hspace{-8pt}
\begin{minipage}[t]{0.58\linewidth}
\vspace{-10pt}

Table~\ref{tab:published-comparison} compares our method with published agent-training baselines. These results are copied from the original papers. We discuss these families of methods in more detail in Section~\ref{sec:related_work}. The comparison is not fully controlled, since the methods differ in training objectives, data construction, and implementation details. However, it is still informative: most methods use similar-scale open backbones such as Qwen2.5-7B or Qwen3-8B, and overlapping results are reported under the same settings, including the 1{,}000-product WebShop setting and the ETO SciWorld split.

Our method is also strong compared with published agent-training results. On ALFWorld, it reaches 89.6\% success, only 1.2 points below the strongest listed result.
\end{minipage}

On WebShop, it reaches 76.8\% success, essentially tied with the strongest listed result of 76.9\%. On SciWorld, it obtains the best listed score, 79.7 versus EMPO2's 75.9. We obtain these results with a modest training budget, with all training runs using a single A100 GPU, whereas strong published RL baselines such as GiGPO report using multiple H100 GPUs for training. This performance also comes with a shorter inference context. \revision{Table~\ref{tab:context-efficiency} reports the measured token savings relative to ReAct-1step and ReAct-full under controlled settings.}

\vspace{-10pt}
\section{Related Work}\label{sec:related_work}

\vspace{-5pt}
\noindent\textbf{Prompting and memory-augmented agents.}
ReAct-style agents maintain task progress by placing prior observations, actions, and reasoning traces in the prompt~\citep{yao2022react}. A large line of memory-augmented agents reduces this burden by storing, summarizing, retrieving, or compressing past interactions~\citep{zhang2025survey,xu2025mem,shu2026remem,fang2025lightmem,zhang2025memgen,liang2025sage,liu2025large}. Prompt-compression methods similarly reduce inference cost by shortening the text context~\citep{jiang2023llmlingua,mu2023learning,liskavets2025prompt}. These methods improve long-horizon behavior, but reusable procedural knowledge remains represented as text that must be selected and reprocessed at inference time. Our work takes a complementary direction: recurring procedure is moved into task-family weights, while the prompt carries only the compact information. \revision{GEPA evolves prompts through execution-based reflection~\citep{agrawal2026gepa}, while ACE maintains an evolving playbook of strategies~\citep{zhang2026agentic}. Both adapt context rather than task-family weights, complementing our policy-learning approach.}

\noindent\textbf{Training language agents.}
Recent work improves LLM agents with expert trajectories, self-generated experience, preference optimization, or reinforcement learning~\citep{shinn2023reflexion,shi2024direct,zhou2024archer,fu2025agentrefine,song2024trial,feng2025group,zhang2025agent,liu2026exploratory,liu2026agentic}. These methods substantially improve task performance, and several also incorporate memories or early experiences into training. However, most preserve a prompting-style execution interface, where the agent still conditions on long instructions, histories, retrieved content, or candidate actions. Our pipeline shares the SFT-then-RL training recipe, but trains the policy to act from a bounded-context input consisting of the current observation, one-step context, and a tracker-rendered state block.

\noindent\textbf{Skill modules and subgoal-aligned rewards.} Some agents accumulate reusable skills as retrievable text or code libraries~\citep{wang2023voyager,liu2024odyssey}. \revision{Agent Workflow Memory represents induced workflows as textual guidance~\citep{wang2024agent}, while SkillWeaver refines website skills into callable APIs~\citep{zheng2025skillweaver}.} We share the goal of reusable skills, but move recurring procedures into lightweight task-family weights, avoiding the longer prompts and slower inference caused by repeated skill-text processing. Parameter-efficient adaptation makes this practical without updating the full backbone~\citep{hu2022lora,han2024parameter}. New workflows are added by training new adapters, limiting interference and catastrophic forgetting across heterogeneous task families, especially for smaller local models. Group-normalized RL avoids training a separate value model~\citep{shao2024deepseekmath}. Our contribution is the feedback design, subgoal rewards are aligned with the tracker fields rendered in the state block, encouraging the module to use compact state during interaction~\citep{sutton1999between,kulkarni2016hierarchical,ng1999policy}.

\vspace{-10pt}
\section{Conclusion and Limitations}\label{sec:conclusion}
\vspace{-5pt}
We presented constant-context skill learning, a context-to-weights framework for recurring LLM-agent workflows. Instead of repeatedly prompting an agent with long skill descriptions and growing histories, we train lightweight task-family modules that internalize reusable procedure while conditioning on the current observation and a compact tracker-rendered state block. Across ALFWorld, WebShop and SciWorld, the resulting agents achieve competitive task performance while substantially reducing inference context compared with ReAct-style prompting. The results suggest that recurring agent behavior can be made both more capable and more context-efficient by moving procedural context from prompts into weights.

The main limitations concern automation, modality, and transfer. First, the state representation and reward rules are task-family specific. We manually define the tracker fields rendered in the state block, and use a strong reasoning model offline to draft deterministic subgoal rewards from the task description and state-block schema. \revision{The LLM-generated WebShop tracker demonstrates that this construction can be automated, but its 65.8\% SR trails the mature tracker at 76.8\%.} This design burden may become easier to automate as agent systems expose \texttt{SKILL.md}-style specifications of procedures, tool constraints, and success conditions~\citep{anthropic_agent_skills_2026}. Compiling such specifications with execution logs into state-block schemas and aligned subgoal rewards is a promising future direction. Second, our evaluation is limited to text-mode environments. Extending the interface to GUI and multimodal agents is a natural next step. Third, per-family modules do not zero-shot transfer to new skills. Adding a workflow requires training a new adapter, although the base model remains fixed and existing skills are preserved. Moreover, each adapter is only $0.5$--$0.7$ GB on disk and only the active one loads at inference, making the storage cost modest in practice.

\newpage

\begingroup
\small
\bibliographystyle{plainnat}
\bibliography{ref}
\endgroup

\newpage

\section{Technical appendices and supplementary material}\label{app:appendix}

\subsection{Training Details and Hyperparameters}
\label{app:training_details}

All three benchmarks provide expert trajectories, which we use directly to construct the SFT data after benchmark-specific filtering. Each successful trajectory is converted into step-level bounded-context inputs paired with the expert action, so the number of SFT samples is larger than the number of trajectories. Table~\ref{tab:training-cost} reports the resulting data scale and training settings for the main Qwen3-8B experiments. The Qwen3-4B\revision{, Qwen3-14B,} and Llama-3.1-8B runs use the same recipe unless otherwise specified. Because Qwen3-8B supports a chat-template thinking mode, we disable it during both training and evaluation (\texttt{enable\_thinking=False}) and use the model as a standard instruction-following language model.  Qwen3-4B and Llama-3.1-8B do not have a thinking mode and are used with their default chat templates.

ALFWorld contains five base task families in our training setup: Pick, Clean, Heat, Cool, and Examine. Pick2 is not trained as a separate module because it does not introduce a new procedure: it consists of two sequential executions of the same Pick-and-Place skill. At evaluation time, we therefore reuse the Pick module twice rather than learning an additional Pick2-specific module.

WebShop is treated as a single task family because all examples share the same shopping workflow. To align with standard WebShop evaluation, we use the 1K-product text environment and evaluate on 500 held-out goals.

For SciWorld, we follow the ETO evaluation protocol, which is the common setup used by recent ScienceWorld agent-training work, including the task-family grouping, official train/dev/test splits, and per-task step limits~\citep{song2024trial}. We keep the family definitions fixed rather than regrouping tasks manually. We train nine task-family modules: find\_X, lifespan, conductivity, measurement, growth, state\_change, power, chemistry, and life\_stages. The first five families have more expert data, with 480, 254, 256, 348, and 145 training variants respectively. Their SFT policies are already near-saturated, so we do not apply RL refinement to them. The remaining four families are lower-data: state\_change has 56 training variants, power has 16, chemistry has 52, and life\_stages has 10. We apply RL refinement to these four families in the main SciWorld setting. Evaluation is performed over the full ETO dev/test sets used in our experiments, containing 194 seen and 211 unseen variants across the nine families.

\begin{table}[t]
  \centering
  \caption{Training and evaluation data scale, hyperparameters, and evaluation protocol for the main Qwen3-8B experiments. SFT samples denote step-level state-action pairs.}
  \label{tab:training-cost}
  \small
  \setlength{\tabcolsep}{8pt}
  \begin{tabular*}{\linewidth}{@{\extracolsep{\fill}}l c c c@{}}
  \toprule
  Hyperparameter & ALFWorld & WebShop & SciWorld \\
  \midrule
  \multicolumn{4}{@{}l}{\textit{Backbone and adapter}} \\
  \quad Base model & \multicolumn{3}{c}{Qwen3-8B (frozen)} \\
  \quad Adapter & \multicolumn{3}{c}{LoRA, rank 64, $\alpha=128$} \\
  \quad Target modules & \multicolumn{3}{c}{q, k, v, o, gate, up, down} \\
  \quad Trainable params & \multicolumn{3}{c}{$\sim$175M (2.09\% of base)} \\
  \quad Precision & \multicolumn{3}{c}{bf16 mixed} \\
  \quad Optimizer & \multicolumn{3}{c}{AdamW} \\
  \midrule
  \multicolumn{4}{@{}l}{\textit{Training data}} \\
  \quad Task-family modules & 5 & 1 & 9 \\
  \quad Expert trajectories & 2{,}628 & 1{,}571 & 1{,}617 \\
  \quad SFT samples & 36{,}702 & 15{,}087 & 37{,}403 \\
  \midrule
  \multicolumn{4}{@{}l}{\textit{Supervised fine-tuning}} \\
  \quad Learning rate & \multicolumn{3}{c}{$2\times 10^{-4}$} \\
  \quad Effective batch & \multicolumn{3}{c}{16} \\
  \quad Epochs & 3 & 3 & 1 \\
  \quad Loss masking & \multicolumn{3}{c}{action tokens only} \\
  \quad LR scheduler & \multicolumn{3}{c}{cosine, warmup ratio 0.1} \\
  \quad LoRA dropout & \multicolumn{3}{c}{0.05} \\
  \quad Max seq length & 1024 & 2048 & 1024 \\
  \midrule
  \multicolumn{4}{@{}l}{\textit{Reinforcement learning}} \\
  \quad Initialization & \multicolumn{3}{c}{SFT adapter} \\
  \quad Reference model & \multicolumn{3}{c}{frozen SFT adapter} \\
  \quad RL epochs & \multicolumn{3}{c}{1} \\
  \quad Rollouts per group $K$ & \multicolumn{3}{c}{4} \\
  \quad Rollout temperature & \multicolumn{3}{c}{0.8} \\
  \quad Rollout top-$p$ & \multicolumn{3}{c}{0.95} \\
  \quad Learning rate & \multicolumn{3}{c}{$5\times 10^{-6}$} \\
  \quad Discount $\gamma$ & \multicolumn{3}{c}{0.98} \\
  \quad KL coefficient $\beta$ & \multicolumn{3}{c}{0.02} \\
  \quad Gradient clip & \multicolumn{3}{c}{1.0} \\
  \quad Train games & 200/family & 200 & low-data families only \\
  \quad Max episode steps & 30 & 15 & ETO per-task limits \\
  \midrule
  \multicolumn{4}{@{}l}{\textit{Evaluation}} \\
  \quad Eval games (seen/unseen) & 140 / 134 & 500 & 194 / 211 \\
  \quad Performance eval & \multicolumn{3}{c}{temperature 0.4, top-$p$ 0.95; seeds 0, 1, 2} \\
  \quad Token-cost eval & \multicolumn{3}{c}{greedy ($T=0$), seed 42} \\
  \quad Max new tokens & \multicolumn{3}{c}{64} \\
  \midrule
  \multicolumn{4}{@{}l}{\textit{Compute}} \\
  \quad Hardware & \multicolumn{3}{c}{1$\times$A100 80GB} \\
  \quad Training seed & \multicolumn{3}{c}{42} \\
  \quad SFT wall-clock & $\sim$173 min & $\sim$143 min & $\sim$90 min \\
  \quad RL wall-clock & $\sim$650 min & $\sim$86 min & $\sim$130 min\\
  \bottomrule
  \end{tabular*}
\end{table}

\subsection{Reward Specification}
\label{app:reward_details}

Table~\ref{tab:reward-spec} summarizes the immediate per-step rewards used during RL. To make reward construction scalable across task families, we draft the rule sets offline using OpenAI GPT-5.5, prompted with the task description and state-block schema. A representative ALFWorld prompt skeleton is shown in Appendix~\ref{app:reward_prompts}. We then implement the proposed rules as fixed deterministic scoring rules. During RL rollouts, no LLM judge or external model is queried. Discounted returns are computed from these rewards with discount factor $\gamma=0.98$.

\begin{table}[t]
\centering
\caption{Reward specification used for RL. Rewards are deterministic rules computed from the environment response, action, and tracker state.}
\label{tab:reward-spec}
\scriptsize
\setlength{\tabcolsep}{3pt}
\begin{tabularx}{\linewidth}{@{}p{1.35cm} p{2.2cm} X@{}}
\toprule
Benchmark & Component & Reward terms \\
\midrule

\multirow{4}{*}{ALFWorld}
& Environment success
& $+3.0$ if the episode terminates successfully. \\
& Progress rewards
& $+1.0$ for advancing the tracked subgoal; $+0.02$ for visiting a new hinted location type during search; $+0.05$ for opening a new hinted container; $+0.2$ for reaching the destination; $+0.2$ for opening the destination receptacle; $+0.5$ for a correct placement action. \\
& Error penalties
& $-0.3$ for invalid actions; $-0.2$ for no-effect actions when admissible actions are unavailable; $-0.2$ for repeated no-progress; $-0.3$ for late-stage regression; $-0.2$ for looking during placement; $-0.3$ for moving to a wrong destination instance; $-0.1$ for revisiting a location; $-0.1$ for reopening a searched container; $-0.15$ for wandering after the search phase. \\
& Step cost
& $-0.01$ per step. \\
\midrule

\multirow{4}{*}{WebShop}
& Environment success
& $+3.0$ for exact successful purchase, i.e., final score $=1.0$. \\
& Progress rewards
& $+0.15$ for selecting a required option; $+0.10$ when all required options are filled and the phase advances to purchase. \\
& Error penalties
& $-0.10$ for selecting a wrong option; $-0.25$ for premature purchase before required options are filled; $-0.10$ for an $A{\rightarrow}B{\rightarrow}A$ action loop; $-0.08$ for clicking an already visited product; $-0.08$ for revisiting the same detail page. \\
& Step cost
& $-0.01$ per step. \\
\midrule

\multirow{4}{*}{SciWorld}
& Environment score
& $\Delta\mathrm{score}/33.0$, so a 100-point task contributes about $+3.0$ in total; additionally $+1.0$ when the episode terminates with score $\ge 100$. \\
& Progress rewards
& Family-specific milestone bonuses tied to the state block, such as focusing or acquiring the target object, reaching the answer room, connecting circuit components, planting and watering seeds, observing state changes, reading recipes, mixing required materials, or focusing the next life stage. Milestone bonuses are one-time and typically range from $+0.08$ to $+2.0$ depending on their position in the procedure. \\
& Error penalties
& Invalid or no-effect actions, repeated actions, oscillations, premature answers, wrong-focus actions, wrong-skill actions, invalid circuit or mixing operations, and task-specific loops. Penalties are typically between $-0.03$ and $-0.25$. \\
& Step cost
& $-0.01$ per step. \\
\bottomrule
\end{tabularx}
\end{table}

\subsection{Data Efficiency of Expert Trajectories}
\label{app:data_efficiency}

\begin{wrapfigure}[14]{r}{0.44\linewidth}
    \centering
    \vspace{-12pt}
    \includegraphics[width=\linewidth]{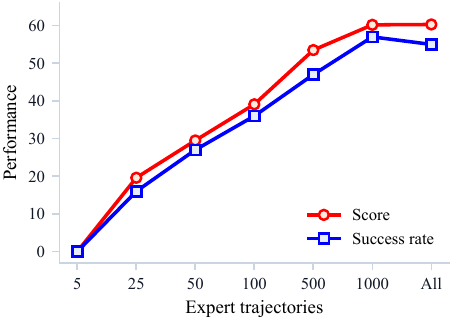}
    \caption{WebShop SFT performance versus expert trajectory count.}
    \label{fig:webshop_traj_sweep}
\end{wrapfigure}
To study how many expert trajectories are needed to learn a recurring workflow, we run a data-efficiency sweep on WebShop with Qwen3-8B. We focus on the SFT stage only, because adding RL would introduce additional interaction budget and make the effect of demonstration count harder to isolate. For each setting, we train the same WebShop skill module with the same hyperparameters as the main experiment, varying only the number of expert trajectories used for SFT. Evaluation is performed on a fixed 100-goal test subset, so these numbers are intended to show trends rather than replace the full 500-goal results in Table~\ref{tab:main-results}.

Figure~\ref{fig:webshop_traj_sweep} shows a clear scaling trend. With only 5 trajectories, the model fails to learn a usable shopping policy. Performance improves steadily from 25 to 1{,}000 trajectories, with success rate increasing from 16.0\% to 57.0\%. Using all available WebShop trajectories gives similar performance to the 1{,}000-trajectory setting, suggesting that SFT begins to saturate at this scale for the fixed 100-goal subset.

\subsection{Qualitative Examples}
\label{app:qualitative_examples}

\begin{figure}[!h]
    \centering
    \includegraphics[width=1\linewidth]{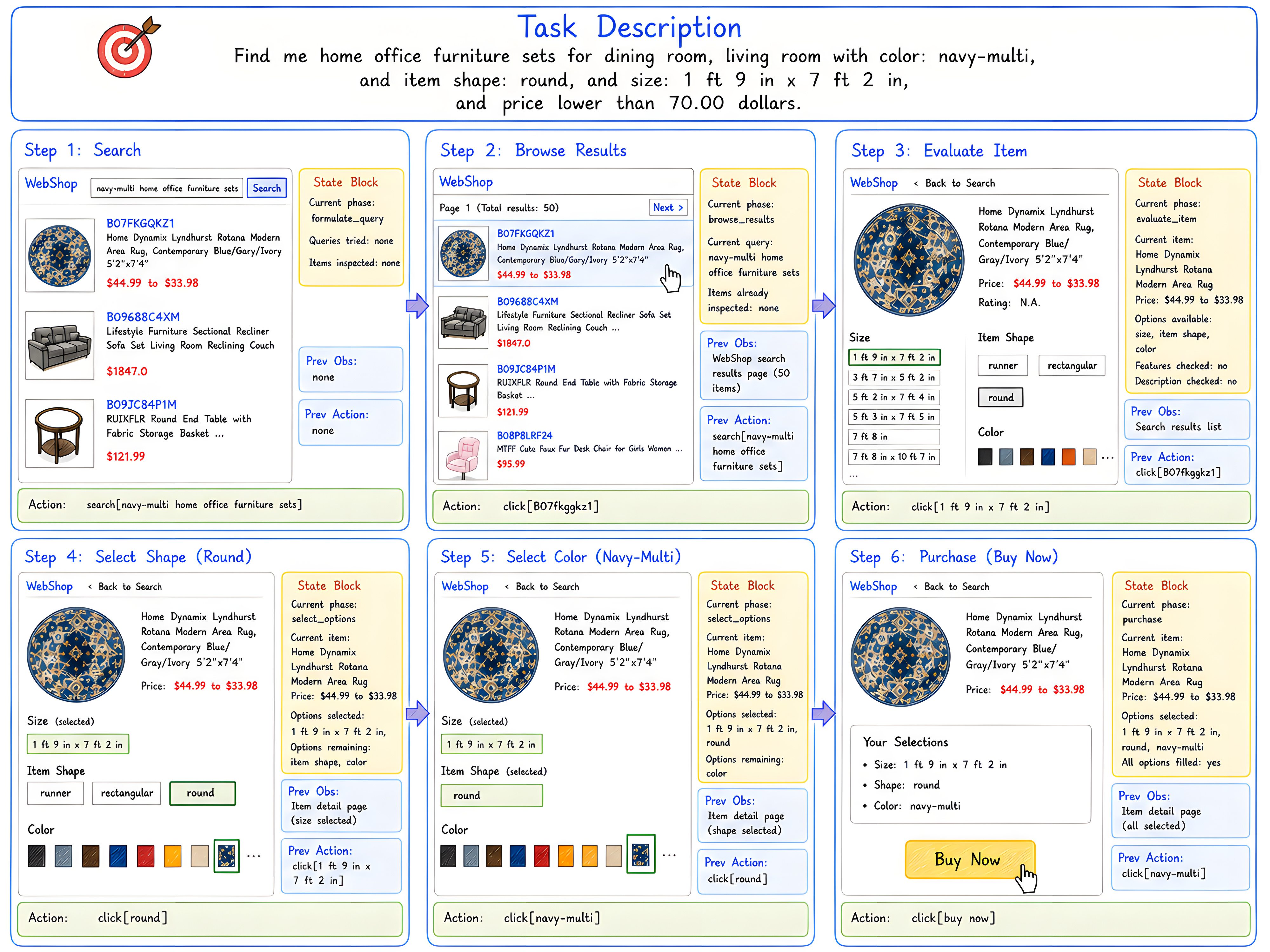}
    \caption{Qualitative WebShop shopping trace with rendered state blocks.}
    \label{fig:webshop_example}
\end{figure}

\begin{figure}[!h]
    \centering
    \includegraphics[width=1\linewidth]{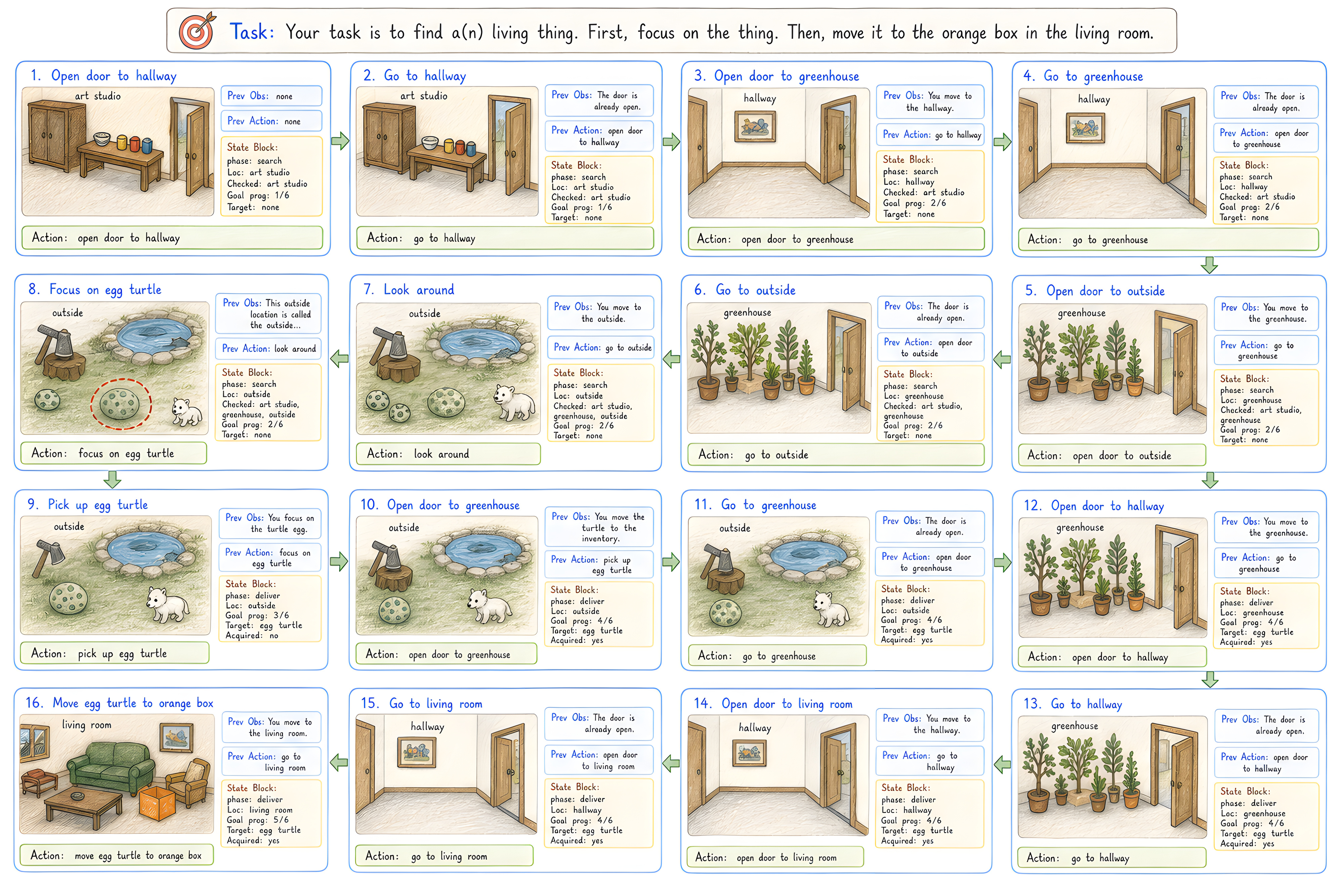}
    \caption{Qualitative SciWorld lab-procedure trace with rendered state blocks.}
    \label{fig:sciworld_example}
\end{figure}

Figures~\ref{fig:webshop_example} and~\ref{fig:sciworld_example} illustrate the same bounded-context interface on WebShop and SciWorld. Each trace shows how the rendered state block, current observation, and one-step context support direct executable action generation without carrying the full episode history.

\subsection{Reward-Drafting Prompt Example}
\label{app:reward_prompts}

To make reward construction reproducible, we release representative
reward-drafting prompts. The prompt below shows the ALFWorld
pick-and-place template. WebShop and SciWorld use the same structure,
with the environment description, action space, and tracker schema
replaced by the corresponding task family. The drafted rules are
manually checked and then implemented as fixed deterministic scoring
rules.

\begin{lstlisting}[
  caption={Reward-drafting prompt skeleton (ALFWorld example).},
  label={lst:reward-prompt}
]
SYSTEM
You are drafting deterministic reward-shaping rules for online RL
refinement of an LLM agent. Your job is to design:
(1) r_progress: subgoal-progress rewards aligned with the state block,
(2) r_error: penalties for actions that block, undo, or violate progress,
(3) r_step: a small step cost to discourage unnecessary actions.

All rules must be computable from the tracker state, executed action,
new observation, and updated tracker state. No LLM judge is allowed at
rollout time.

TASK FAMILY
ALFWorld pick-and-place: put a target object into a target receptacle.

ACTION SPACE
go to X; take X from Y; put X in/on Y; open X; close X; examine X; look.

STATE-BLOCK SCHEMA
target_object       : object to manipulate
destination         : target receptacle
holding             : currently held object or None
checked_locations   : visited receptacles
searched_containers : opened containers
current_subgoal     : find_object | take_object | reach_dest |
                      place_object | done

TRACKER TRANSITION EXAMPLES
- find_object -> take_object when the target object becomes visible.
- take_object -> reach_dest when holding changes from None to target_object.
- reach_dest -> place_object when the agent reaches destination.
- place_object -> done when the object is placed in the destination.
- No progress if the same location is revisited without new evidence.
- Regression if the agent leaves the destination during place_object.

DESIGN INSTRUCTION
Design reward-shaping terms over the above state-block fields and
tracker transitions. The total reward will be:

    r_t = r_env + r_progress - r_error - r_step

where r_env is fixed by the environment. Return a Python-style
dictionary reward_terms.

Rules:
- r_progress should reward tracker-state changes that advance the
  current subgoal or make required state-block fields closer to done.
- r_error should penalize invalid actions, repeated no-progress actions,
  premature placement, wrong-destination behavior, and subgoal regression.
- r_step should be a small positive penalty magnitude applied each step
  to discourage unnecessary wandering or browsing.
- Every trigger must explicitly reference tracker fields, actions,
  observations, or tracker-state changes.
\end{lstlisting}

\begin{revisionblock}
\subsection{Generalization Probe Protocol}\label{app:generalization_protocol}
We evaluate the Qwen3-8B SFT+RL adapters used in Table~\ref{tab:main-results}, without additional training or checkpoint selection. WebShop uses environment seed 2, all 500 standard test goals unless filtered below, and inference seeds 0, 1, and 2. ALFWorld uses all 134 official unseen games with the same three inference seeds. Decoding follows the main performance protocol: temperature 0.4 and top-$p$ 0.95. Each probe is also evaluated with the original frozen Qwen3-8B model, with the same tracker, prompt, parser, task instances, and decoding settings.

\paragraph{Instruction paraphrasing.}
GPT-5.4 rewrites each instruction under a fixed JSON output schema, preserving constraint tokens such as attribute values, sizes, quantities, and price inequalities. The response includes an explicit constraint-preservation check, and a random sample is manually checked. Only the instruction shown to the policy is replaced: the environment goal, reward computation, and tracker parsing use the original instruction. The probe therefore tests the policy's response to wording changes, rather than robustness of the tracker parser to paraphrases. The construction is applied to all 500 WebShop test instructions and all 134 ALFWorld unseen instructions.

\paragraph{Unseen constraint combinations.}
We extract constraint sets from the 1,571 WebShop SFT training goals and retain test goals whose complete joint constraint set does not occur in the SFT goals, although each atomic constraint does. This yields 446 test goals. No LLM is used for this filtering. Because this is a subset of the original 500-goal evaluation, its absolute success rate should not be interpreted as a paired difference from the full-set result. The probe tests combinations absent from the SFT goals, without establishing absence from the backbone's pretraining data or subsequent RL experience.

\paragraph{State-block reserialization.}
We retain every WebShop tracker field and value but render the block as sorted JSON, changing its ordering and delimiters. The tracker update rules, environment, and policy weights are unchanged. This isolates sensitivity to the serialization of the supplied state.

Table~\ref{tab:probe_scores} supplements the success rates in Table~\ref{tab:generalization} with WebShop scores. The frozen base is weak under both the standard and perturbed interfaces. This isolates a substantial contribution from adaptation, while the remaining paraphrase drop shows that the learned policy is still sensitive to some changes in wording. None of these probes changes the environment dynamics or establishes transfer to a new task family.
\end{revisionblock}

\begin{table}[h]
\color{revisionblue}
\captionsetup{labelfont={color=revisionblue},textfont={color=revisionblue}}
\centering
\caption{WebShop scores for the generalization probes (mean$\pm$std over three inference seeds). Success rates and sample sizes appear in Table~\ref{tab:generalization}.}
\label{tab:probe_scores}
\small
\begin{tabular*}{\linewidth}{@{\extracolsep{\fill}}l c c@{}}
\toprule
Input condition & Frozen-base score & SFT+RL score \\
\midrule
Original & $27.3\pm0.9$ & $84.0\pm0.0$ \\
Paraphrased instruction & $28.2\pm0.8$ & $82.7\pm0.5$ \\
Unseen constraint combination & $27.3\pm0.8$ & $84.9\pm0.3$ \\
State block as sorted JSON & $17.4\pm0.6$ & $83.3\pm0.4$ \\
\bottomrule
\end{tabular*}
\end{table}

\begin{revisionblock}
\subsection{Frozen-Base Controls}\label{app:frozen_base_controls}
To isolate the contribution of the learned adapter under the supplied state interface, we evaluate the original frozen Qwen3-8B model without loading any LoRA. Task instances, splits, tracker, state block, prompt template, action parser, environment, decoding parameters, and inference seeds are otherwise unchanged. Table~\ref{tab:frozen_base_controls} compares the standard evaluations with the original SFT and SFT+RL results. All values are mean$\pm$std over the three inference seeds of the main performance protocol.
\end{revisionblock}

\begin{table}[!ht]
\color{revisionblue}
\captionsetup{labelfont={color=revisionblue},textfont={color=revisionblue}}
\centering
\caption{Qwen3-8B with the same state interface before and after adapter training. Entries give Score / SR (\%); ALFWorld reports SR only.}
\label{tab:frozen_base_controls}
\footnotesize
\setlength{\tabcolsep}{4pt}
\begin{tabular*}{\linewidth}{@{\extracolsep{\fill}}l l c c c@{}}
\toprule
Benchmark & Split & Frozen base & SFT & SFT+RL \\
\midrule
ALFWorld & Seen & $0.0\pm0.0$ & $62.4\pm2.5$ & $83.6\pm1.8$ \\
& Unseen & $0.0\pm0.0$ & $61.7\pm2.2$ & $89.6\pm0.4$ \\
WebShop & Test & $27.3\pm0.9$ / $22.5\pm0.8$ & $69.5\pm1.7$ / $62.2\pm1.2$ & $84.0\pm0.0$ / $76.8\pm0.2$ \\
SciWorld & Seen & $11.6\pm1.2$ / $4.8\pm0.3$ & $65.7\pm1.6$ / $58.2\pm1.5$ & $72.8\pm2.7$ / $62.9\pm2.2$ \\
& Unseen & $12.3\pm0.3$ / $4.9\pm0.7$ & $61.5\pm0.9$ / $52.0\pm0.7$ & $79.7\pm0.2$ / $66.4\pm0.5$ \\
\bottomrule
\end{tabular*}
\end{table}

\begin{revisionblock}
Table~\ref{tab:frozen_probe_controls} extends this control to the generalization probes. WebShop score counterparts are reported in Table~\ref{tab:probe_scores}. The base model reaches only 13.5--22.6\% SR under the WebShop perturbations and 0.0\% on ALFWorld, despite receiving the same tracker and state information. The adapted policy's performance and robustness therefore exceed those of the unadapted backbone. These controls concern the tested task families and do not establish transfer to new environments.
\end{revisionblock}

\begin{table}[!ht]
\color{revisionblue}
\captionsetup{labelfont={color=revisionblue},textfont={color=revisionblue}}
\centering
\caption{Frozen-base controls for the generalization probes with Qwen3-8B (mean$\pm$std over three inference seeds). The frozen base and trained policy receive the same state interface. $N$ is the number of goals per seed; SR is in percent.}
\label{tab:frozen_probe_controls}
\small
\setlength{\tabcolsep}{4pt}
\begin{tabularx}{\linewidth}{@{}l X r c c@{}}
\toprule
Benchmark & Input condition & $N$ & Frozen-base SR & SFT+RL SR \\
\midrule
WebShop & Original & 500 & $22.5\pm0.8$ & $76.8\pm0.2$ \\
& Paraphrased instruction & 500 & $22.5\pm0.6$ & $72.2\pm0.7$ \\
& Unseen constraint combination & 446 & $22.6\pm0.9$ & $75.8\pm0.6$ \\
& State block as sorted JSON & 500 & $13.5\pm0.2$ & $75.7\pm0.8$ \\
\midrule
ALFWorld & Original unseen & 134 & $0.0\pm0.0$ & $89.6\pm0.4$ \\
& Paraphrased unseen & 134 & $0.0\pm0.0$ & $89.1\pm1.1$ \\
\bottomrule
\end{tabularx}
\end{table}

\begin{revisionblock}
\subsection{LLM-Generated WebShop Tracker}\label{app:auto_tracker}
We test whether the state interface can be constructed with an LLM by replacing the authors' WebShop tracker with an LLM-generated implementation. GPT-5.4 receives the public text-mode environment contract, aggregate coverage and selected examples from the 1,571 public SFT trajectories, and the public option-group vocabulary. The authors' manual tracker and evaluation goals 0--499 are excluded from generation. The model produces a state schema, deterministic parser and update logic, bounded JSON renderer, prompt builder, action extractor, and dense reward rules. Integration repairs are also LLM-produced, with deterministic trajectory replay supplying validation signals.

Validation replays all 15,111 states from the training trajectories, including 6,523 item-page states. It checks markup removal, product-title extraction, separation of titles from selectable options, and tracker execution. The finalized implementation records zero failures on these checks and makes no LLM, API, or network call during training or inference. These checks cover the observed replay states; they do not guarantee correct tracking on every possible future state.

We train a Qwen3-8B LoRA for three SFT epochs, then refine it on 200 training games with four rollouts per game. The run uses one A100 80GB and training seed 42. Both checkpoints are evaluated on all 500 held-out goals with greedy decoding, seed 42, and a 15-step limit. Table~\ref{tab:auto_tracker} reports this single-run proof of concept alongside the original manual-tracker results. The two implementations use the same source expert corpus, with tracker-specific preprocessing; the manual-tracker reference is a three-inference-seed mean, so this is a descriptive comparison.
\end{revisionblock}

\begin{table}[h]
\color{revisionblue}
\captionsetup{labelfont={color=revisionblue},textfont={color=revisionblue}}
\centering
\caption{WebShop performance with manual and LLM-generated trackers. Manual-tracker numbers are the original three-seed means; generated-tracker numbers are single greedy evaluations on 500 goals.}
\label{tab:auto_tracker}
\small
\begin{tabular*}{\linewidth}{@{\extracolsep{\fill}}l c c@{}}
\toprule
Tracker and training stage & Score & SR (\%) \\
\midrule
Manual tracker, SFT & 69.5 & 62.2 \\
Manual tracker, SFT+RL & 84.0 & 76.8 \\
LLM-generated tracker, SFT & 72.6 & 61.2 \\
LLM-generated tracker, SFT+RL & 75.0 & 65.8 \\
\bottomrule
\end{tabular*}
\end{table}

\begin{revisionblock}
RL improves success by 4.6 percentage points with the generated tracker. The final 65.8\% SR shows that an LLM-generated interface can support the complete training pipeline, but remains 11.0 points below the mature tracker reference. This is evidence for feasible offline construction followed by validation, rather than parity with manual engineering or autonomous discovery of useful state representations in arbitrary domains.

\paragraph{Generation request.}
The generation prompt supplies the environment's page types, action syntax, option-group vocabulary, trajectory-coverage summary, and representative transitions. It requests Python tracker code and tests, the state schema, deterministic reward rules, assumptions, and a review checklist under a fixed output schema. The core requirements are summarized below. The environment-specific examples are supplied separately.
\end{revisionblock}

\begingroup
\captionsetup[lstlisting]{labelfont={color=revisionblue},textfont={color=revisionblue}}
\begin{lstlisting}[
  basicstyle=\ttfamily\footnotesize\color{revisionblue},
  caption={\revision{Condensed requirements supplied for offline tracker generation.}},
  label={lst:auto_tracker_request}
]
INPUTS
Public WebShop text-interface contract, option-group vocabulary,
training-trajectory coverage, and representative transitions.
Do not use the authors' tracker or held-out evaluation goals.

OUTPUTS
State schema; deterministic Python parser and update logic;
bounded JSON renderer; prompt builder and action extractor;
dense reward rules; tests; assumptions and review checklist.

CONSTRAINTS
Use observable actions, pages, and tracked progress.
No model, API, or network calls in the generated runtime.
Compute dense rewards from observable transitions and terminal score.
Validate against training-trajectory replay and repair failures.
\end{lstlisting}
\endgroup

\begin{revisionblock}
\subsection{A Unified LoRA Across Three Benchmarks}\label{app:unified_cross_benchmark}
To examine whether per-family modules are an architectural requirement, we train one Qwen3-8B LoRA jointly on SFT examples from ALFWorld, WebShop, and SciWorld, using benchmark-balanced sampling. The same adapter is evaluated through each benchmark's native environment and tracker. This experiment changes parameter sharing while retaining the benchmark-specific state interfaces; it does not test execution in an unseen environment. The adapter is SFT-only, with no RL refinement, and the results are single evaluation runs. The SciWorld run uses alias-only action compilation with its action guard disabled.

Table~\ref{tab:unified_cross_benchmark} compares this adapter with the independent Qwen3-8B SFT modules from Table~\ref{tab:main-results}. WebShop remains close to the independent-module reference and SciWorld improves on both splits. ALFWorld unseen success falls by 6.5 points, indicating interference despite benchmark-balanced sampling. The unified numbers are single runs and the independent-module numbers are inference-seed means, so these differences are descriptive, rather than significance tests.
\end{revisionblock}

\begin{table}[h]
\color{revisionblue}
\captionsetup{labelfont={color=revisionblue},textfont={color=revisionblue}}
\centering
\caption{Independent and cross-benchmark unified Qwen3-8B SFT adapters. Entries give Score / SR (\%); ALFWorld reports SR only. Independent values are mean$\pm$std; unified values are single-run results. Neither column includes RL refinement.}
\label{tab:unified_cross_benchmark}
\small
\setlength{\tabcolsep}{5pt}
\begin{tabular*}{\linewidth}{@{\extracolsep{\fill}}l l c c@{}}
\toprule
Benchmark & Split & Independent SFT & Unified SFT \\
\midrule
ALFWorld & Seen & $62.4\pm2.5$ & 65.7 \\
& Unseen & $61.7\pm2.2$ & 55.2 \\
WebShop & Test & $69.5\pm1.7$ / $62.2\pm1.2$ & 70.2 / 60.8 \\
SciWorld & Seen/dev & $65.7\pm1.6$ / $58.2\pm1.5$ & 70.2 / 64.4 \\
& Unseen/test & $61.5\pm0.9$ / $52.0\pm0.7$ & 71.6 / 56.4 \\
\bottomrule
\end{tabular*}
\end{table}

\begin{revisionblock}
Together with the within-ALFWorld unified module in Section~\ref{sec:task_performance}, this result shows that one adapter can support multiple task families. Per-family deployment retains a practical benefit: adding a workflow updates only its new adapter, leaving existing deployed adapters fixed. The unified result demonstrates joint learning across heterogeneous interfaces, while the remaining ALFWorld interference motivates retaining specialization as a deployment option.
\end{revisionblock}

\end{document}